\documentclass[final]{cvpr}

\usepackage{times}
\usepackage{epsfig}
\usepackage{graphicx}
\usepackage{amsmath}
\usepackage{amssymb}
\usepackage{paralist}
\usepackage{csquotes}
\usepackage{rotating}
\usepackage{multirow}

\usepackage[pagebackref=true,breaklinks=true,letterpaper=true,colorlinks,bookmarks=false]{hyperref}

\def\cvprPaperID{2102} 

\begin{document}




\title{Uncertainty-Aware Deep Calibrated Salient Object Detection}


\author{
Jing Zhang$^{1,3,4}$\quad
Yuchao Dai$^{2}$\thanks{Corresponding author: Yuchao Dai \emph{(daiyuchao@gmail.com)}}\quad
Xin Yu$^{^5}$ \quad 
Mehrtash Harandi$^{6}$\quad 
Nick Barnes$^{1}$\quad 
Richard Hartley$^1$  \\
$^1$ Australian National University \quad
$^2$ Northwestern Polytechnical University 
$^3$ ACRV \quad \\
$^4$ Data61 \quad 
$^5$ ReLER, University of Technology Sydney \quad 
$^6$ Monash University \quad 
\\
}

\def\JZ#1{{\color{red}{\bf [Jing:} {\it{#1}}{\bf ]}}}
\def\YD#1{{\color{blue}{\bf [Yuchao:} {\it{#1}}{\bf ]}}}
\def\XY#1{{\color{blue}{\bf [Xin:} {\it{#1}}{\bf ]}}}
\def\NB#1{{\color{green}{\bf [Nick:} 
{\it{#1}}{\bf ]}}}
\def\MH#1{{\color{violet}{\bf [MH:}{\it{#1}}{\bf ]}}}

\maketitle

\begin{abstract}
Existing deep neural network based salient object detection (SOD) methods mainly focus on pursuing high network accuracy. However, those methods overlook the gap between network accuracy and prediction confidence, known as the confidence uncalibration problem. Thus, state-of-the-art SOD networks are prone to be overconfident.
In other words, the predicted confidence of the networks does not reflect the real probability of correctness of salient object detection, which significantly hinder their real-world applicability.
In this paper, we introduce an uncertainty-aware deep SOD network, and propose two strategies from different perspectives to prevent deep SOD networks from being overconfident. 
The first strategy, namely \textbf{Boundary Distribution Smoothing (BDS)}, generates continuous labels by smoothing the original binary ground-truth with respect to pixel-wise uncertainty.
The second strategy, namely \textbf{Uncertainty-Aware Temperature Scaling (UATS)}, exploits a relaxed Sigmoid function during both training and testing
with spatially-variant temperature scaling to produce softened output.
Both strategies can be incorporated into existing deep SOD networks with minimal efforts.
Moreover, we propose a new saliency evaluation metric, namely \textbf{dense calibration measure} $\mathcal{C}$,
to measure how the model is calibrated on a given dataset. 
Extensive experimental results on seven
benchmark datasets demonstrate that our solutions can not only better calibrate SOD models, but also improve the network accuracy.
\end{abstract}

\section{Introduction}

There has been profound progress in visual salient object detection (SOD) with the help of deep convolutional neural networks \cite{VGG, ResHe2015}, especially fully convolutional neural networks \cite{FCN,badrinarayanan2015segnet}. 
State-of-the-art SOD methods mainly focus on pursuing high network accuracy by exploiting different backbone deep networks (\eg, VGG-Net \cite{VGG}, ResNet \cite{ResHe2015}), incorporating different prior knowledge \cite{picanet,BASNet_Sal,Background-Detection:CVPR-2014} or learning from weak supervision \cite{imagesaliency,Guanbin_weaksalAAAI, MSW_Sal}.

\begin{figure}[!t]
   \begin{center}
   \begin{tabular}{ c@{ } c@{ }}
   {\includegraphics[width=0.495\linewidth]{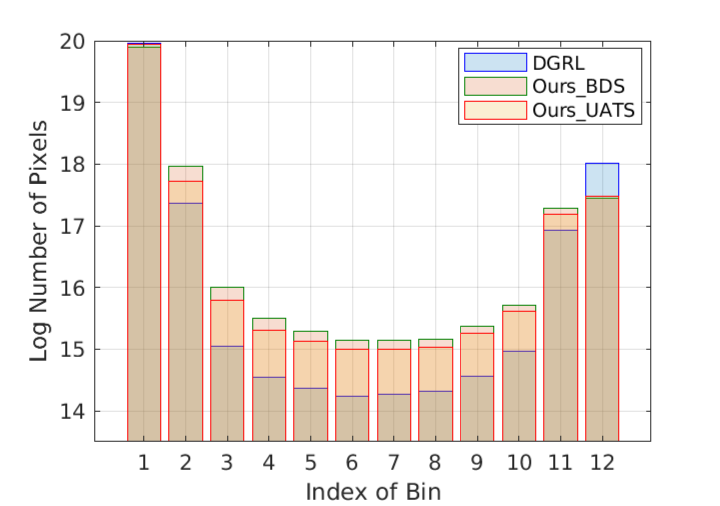}} & {\includegraphics[width=0.495\linewidth]{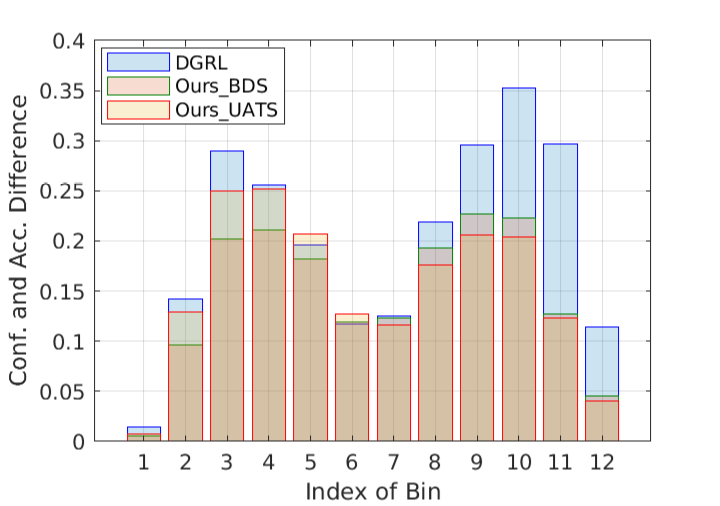}}\\
   \footnotesize{(a)} & \footnotesize{(b)} \\
   \end{tabular}
   \end{center}
   \vspace{-3mm}
\caption{Calibration comparisons on DUTS testing dataset \cite{imagesaliency}. (a) number of samples in each bin for DGRL \cite{Wang_2018_CVPR}, our methods: ``Ours\_BDS'' and ``Ours\_UATS''. (b) shows confidence and accuracy difference for DGRL and our methods.}


   \label{fig:calib_compare}
\end{figure}

SOD \cite{CPD_Sal,BASNet_Sal,picanet,Wang_2018_CVPR} is generally formulated as a binary classification problem, where each pixel is classified as either the salient foreground (1) or the background (0) by a deep
network. 
The Binary Cross-Entropy Loss (BCE) is
employed to optimize network parameters with the ground-truth binary supervision. The network \emph{accuracy} measures the extent to which the network prediction is consistent with the ground-truth.
The network output is normalized by Softmax or the Sigmoid function, and it is termed the \emph{confidence},
representing how the model trusts its prediction.

In this paper, we would like to raise a natural question that \emph{whether confidence of SOD network
is consistent with the network accuracy?}
This question is referred to the problem of confidence calibration \cite{Guo2017OnCO}, where Guo~\etal \cite{Guo2017OnCO} discovered that modern deep neural networks are poorly calibrated, \ie, the confidence and the accuracy are not consistent. Their conclusion is based on the image-level classification problem. In this paper, we extend the analysis of confidence calibration for single label classification to its dense prediction counterpart, salient object detection in particular.

Taking one state-of-the-art SOD model DGRL \cite{Wang_2018_CVPR} as an example, we investigate its confidence calibration issue on DUTS testing dataset \cite{imagesaliency} and show the results in Fig.~\ref{fig:calib_compare}, where \enquote{Oues\_BDS} and \enquote{Ours\_UATS} are models using the BDS and UATS strategies respectively.
Specifically, we group saliency prediction of DGRL \cite{Wang_2018_CVPR} and our method on the above mentioned two testing datasets to 12 bins and compute both network confidence and model accuracy in each bin. A perfectly calibrated model should have \enquote{confidence=accuracy} in all the bins. As reported in Fig.~\ref{fig:calib_compare}, state-of-the-art SOD model DGRL is not well calibrated as evidenced by the gap between confidence and accuracy. By contrast, our proposed methods greatly decrease the gap and therefore are well calibrated.


For SOD, as an important intermediate step in various vision systems \cite{Segmentation_saliency, Xu2015show,SemanticLabel}, we argue that consistent predictions of confidence and accuracy are more desirable than overconfident predictions.
In this paper, we address the confidence calibration issue from uncertainty estimation perspective for the task of deep SOD,
and propose two strategies from different perspectives to prevent the SOD networks from becoming overconfident:
\textbf{1}) relax the supervision signals, and \textbf{2}) soften the model prediction.

\begin{figure}[!t]
   \begin{center}
   \begin{tabular}{ c@{ } c@{ } c@{ }}
   {\includegraphics[width=0.31\linewidth]{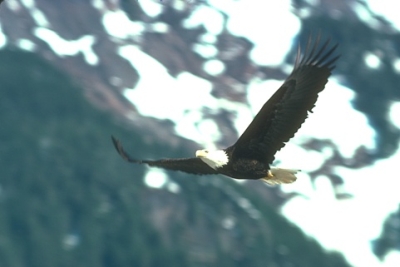}} & {\includegraphics[width=0.31\linewidth]{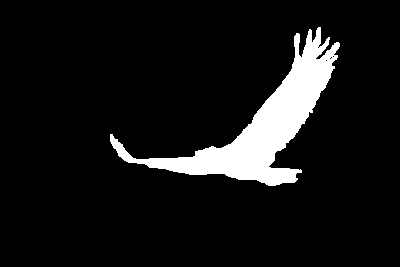}} & {\includegraphics[width=0.31\linewidth]{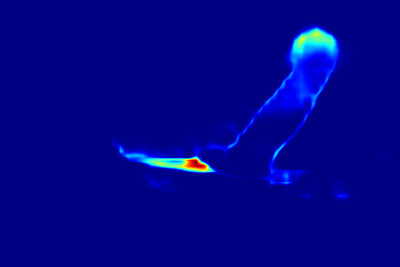}}\\
   \footnotesize{Image} & \footnotesize{Ground-Truth} & \footnotesize{Uncertainty} \\
   \end{tabular}
   \end{center}
   \vspace{-1mm}
\caption{Visualization of the uncertainty map for a given image.
}\vspace{-1em}
   \label{fig:uncertainty_map}
\end{figure}

Firstly, we note that pixels across an entire image are not equal in terms of prediction confidence \cite{li2017not}, which is also consistent with human visual perception. In particular, humans tend to make mistakes around edges. Fig.~\ref{fig:uncertainty_map} shows the uncertainty map of a given image,
which clearly shows that most uncertainty pixels occur along objects edges. We take image uncertainty into account, and
propose Boundary Distribution Smoothing (BDS), a strategy to replace the hard binary supervision with continuous soft labels based on an uncertainty map.


Secondly, we re-map the network confidence in a nonlinear way by taking the spatially-variant nature of uncertainty in dense labelling tasks into account. Our second strategy, namely Uncertainty-aware Temperature Scaling (UATS), uses a relaxed Sigmoid function
to produce softened output. Specifically, for a pixel, we assign high temperature if it belongs to a high uncertainty region, and a low temperature otherwise based on a learned uncertainty map.
Existing temperature scaling (TS) related model calibration technique \cite{Guo2017OnCO} treats TS as a post-processing method and uses a fixed temperature for the entire dataset. Differently,
we produce uncertainty map while training the network, and produce spatial-varying temperature to achieve a well-calibrated and high-accurate model.

Furthermore, we propose dense calibration measure $\mathcal{C}$ as a new evaluation metric to quantitatively measure how each SOD model is calibrated on a specific dataset.
We believe that our work is the first attempt to learn dense prediction models from a model calibration perspective. Both strategies can be incorporated in the training procedure of existing deep SOD networks with minimal efforts.
Experimental results on seven SOD benchmark datasets demonstrate the effectiveness of our proposed strategies. 

Our main contributions can be summarized as:
\vspace{-2mm}
\begin{itemize}
    \item 
    We propose two strategies to prevent the SOD network from overconfident, namely boundary distribution smoothness and uncertainty-aware temperature scaling by taking image uncertainty into account.
    \vspace{-2mm}
    \item We introduce dense calibration measure as an evaluation metric to quantitatively measure how the salienct object detection model is calibrated.
    \vspace{-2mm}
    \item Experimental results on seven SOD benchmark dataset illustrate that our proposed strategies can lead to a well-calibrated model of high accuracy.
\end{itemize}

\section{Related Work}
\label{related_work}
We briefly review deep SOD models, and discuss efforts in addressing the network confidence calibration issue.


\subsection{Deep Salient Object Detection Models}
Depending on how pixel-wise human annotations are used, existing deep SOD models can be classified into three categories: fully supervised models, weakly supervised models and unsupervised models.

\vspace{1mm}
\noindent\textbf{Fully supervised models} \cite{BASNet_Sal,CPD_Sal,AFNet_Sal,MSNet_Sal,ras_reverse, picanet,Wang_2018_CVPR, Amulet,NLDF,wangiccv17,ChengCVPR17} use clean pixel-wise human-annotations as supervision signals.
Wu \etal \cite{CPD_Sal} proposed a cascaded partial decoder framework to achieve fast and high resolution SOD. 
Qin \etal \cite{BASNet_Sal} included a hybrid loss for boundary-aware saliency detection.
Shen \etal \cite{ras_reverse} introduced a reverse attention module to refine a saliency map gradually in a top-down manner.
Liu \etal \cite{picanet} presented pixel-wise contextual attention to effectively select informative context for each pixel. 
With the same purpose, Wang \etal \cite{Wang_2018_CVPR} integrated a global recurrent localization network with a local boundary refinement network to learn both global and local context information. 

\vspace{1mm}
\noindent \textbf{Weakly supervised models} learn saliency from low-cost annotations.
Zeng \etal \cite{MSW_Sal} trained a salient object detection model with diverse weak supervision sources, including category labels, captions, and unlabelled data.
Wang \etal \cite{imagesaliency} introduced a foreground inference network (FIN) within a two-stage learning framework to learn saliency from image-level supervision.
Li \etal \cite{Guanbin_weaksalAAAI} took an coarse saliency map from unsupervised saliency method as initial guess, and then iteratively update it
with image level labels.

\vspace{1mm}
\noindent \textbf{Unsupervised models} \cite{Zhang_2017_ICCV, Zhang_2018_CVPR} do not require large-scale manually labelled annotations. In general, those methods exploit multiple subjective or unreliable annotations. 
Zhang \etal \cite{Zhang_2017_ICCV} learned a deep model, driven by fusing outputs of unsupervised methods heuristically to predict saliency maps.
Given noisy saliency maps from multiple conventional handcrafted feature based methods, \cite{Zhang_2018_CVPR} addressed SOD as the problem of learning from crowds, and updated the network parameters and noise module parameters in an alternating manner. 


\subsection{Confidence Calibration of Deep Networks}
Guo \etal \cite{Guo2017OnCO} discovered that modern deep neural networks are prone to producing overconfident predictions. To remedy this issue, \cite{VAT_tpami,distribution_smooth} generate adversarial samples~\cite{Goodfellow2015ExplainingAH} to enlarge diversity of training dataset. More recently, Kuamr \etal \cite{pmlr-v80-kumar18a} introduced trainable model calibration error as a regularization term to their objective function.
\cite{can_you_trust_nips} investigated the effect of dataset shift on accuracy and calibration by comparison of existing model calibration methods. 
Here, we mainly focus on two related directions for confidence calibration: label relaxation \cite{rethink_inception} and temperature scaling \cite{Guo2017OnCO}.

\noindent\textbf{Label Relaxation} aims at relaxing the supervision signals, thus generating smoothing labels \cite{rethink_inception} or disturbed labels \cite{disturblabel}.
Thulasidasan \etal \cite{On_Mixup_Training_Nips19} discovered that mixup-training \cite{mixup} with label smoothing can significantly improve model calibration.
To obtain a more robust and generative model, Xie \etal \cite{disturblabel} randomly replaced a part of labels as incorrect values in each iteration. 
Griffiths \etal \cite{human_uncertainty_robust_iccv} presented a soft-label dataset (CIFAR10H) aiming at reflecting human perceptual uncertainty by providing label distribution across categories instead of a hard one-hot label.

\noindent\textbf{Temperature Scaling} focuses on modifying the model prediction.
A simple temperature scaling method was proposed in \cite{Guo2017OnCO} to deal with the network overconfidence issue as a post-processing method. 
Neumann \etal \cite{relaxed_softmax} proposed a relaxed Softmax layer based on sample-dependent temperature.
Hinton \etal \cite{distilling_knowledge} proposed to raise the temperature of the final Softmax until the model produced a suitably soft set of targets.



Although model calibration has been studied for the image-level classification task, we would like to emphasize that no such study exists in saliency prediction.
Meanwhile, adapting existing model calibration strategies to SOD networks is not straight-forward. Firstly, most existing work on model calibration focus on image-level classification problem \cite{Guo2017OnCO}, where there exists no context relationships inside each sample. For SOD, context inside the image plays a key role. Secondly, different from image-level classification, category information is not available in saliency prediction model, thus introducing soft-label dataset like \cite{human_uncertainty_robust_iccv} to reflect category relationship does not work.
In this paper, we present two strategies to produce uncertain-aware deep calibrated salient object detection model, and we will discussed both solutions in details in Section \ref{our_framework}.

\section{Calibrated Salient Object Detection}
\label{our_framework}
As to the nature of salient object detection and the smooth changes of human attention towards a natural images, it is more reasonable to produce a confidence map to indicate the uncertainty of saliency prediction. This motivates us to study the uncalibrated phenomenon in SOD rather than attaining an accuracy orientated binary classification based deep model.
Inspired by recent advances in confidence calibration~\cite{human_uncertainty_robust_iccv,Guo2017OnCO,distilling_knowledge,NIPS2018_7471} and adversarial samples~\cite{rethink_inception, mixup},
we extend confidence calibration from the single label classification problem to its dense labelling counterpart, in particular salient object detection. We present two strategies, namely Boundary Distribution Smoothing (BDS) and Uncertainty-aware Temperature Scaling (UATS) to alleviate the over-confidence problem in deep SOD models. We further propose dense calibration measure as an evaluation metric to measure how the network is calibrated.




\subsection{Rethinking Deep Salient Object Detection}

Let $D=\{\pmb{x}_i,\pmb{y}_i\}_{i=1}^N$ be a training set, where $\pmb{x}_i \in \mathbb{R}^{h \times w \times 3}$ is an image and $\pmb{y}_i \in \mathbb{R}^{h \times w}$ is the corresponding ground-truth saliency map. Pixels with value 0 in $\pmb{y}_i$ encode the background of image $\pmb{x}_i$ while the salient objects are marked with 1.

Conventional deep solutions approach SOD as a dense binary classification problem. More specifically, a deep SOD model is learned by minimizing the empirical risk:
\begin{equation}
\label{erm}
    \mathcal{R} = \frac{1}{N}\sum_{i=1}^N\sum_{(u,v)} \ell(\pmb{s}_i^{(u,v)},\pmb{y}_i^{(u,v)}),
\end{equation}
where $\pmb{s}_i=p(\pmb{y}_i|\pmb{x}_i,\Theta)$ is normalized output with $\Theta$ as the parameter set of the network, $(u,v)$ denotes coordinates of pixels in an image. Generally, the loss $\ell$ is the binary cross-entropy loss defined as:
\begin{align}
\label{raw_image_loss}
\ell(s, y) = -\big( y\, \log(s) + (1-y)\,\log(1-s)\big) ~,
\end{align}
where $s=\pmb{s}_i^{(u,v)}$ and $y=\pmb{y}_i^{(u,v)}$, correspond to network prediction and ground-truth at coordinate $(u,v)$.

The empirical risk in Eq.~\eqref{erm} is minimized by using variants of stochastic gradient descent.
The minimum of the above loss function is achieved with peaked predictions, where $s$ approaches 1 for the foreground (salient) objects and 0 otherwise as shown in Fig.~\ref{fig:calib_compare} (a).
The main consequence of the peaked prediction is that the learned model could even output high confidence for incorrect prediction (low accuracy). One of the solution to cope with the overconfidence problem is uncertainty modeling \cite{kendall2017uncertainties,kendall2017bayesian,Kendall_2018_CVPR}.

\begin{figure}[h]
   \begin{center}
   \begin{tabular}{ c@{ } c@{ } c@{ } c@{ } c@{ }}
   {\includegraphics[width=0.185\linewidth]{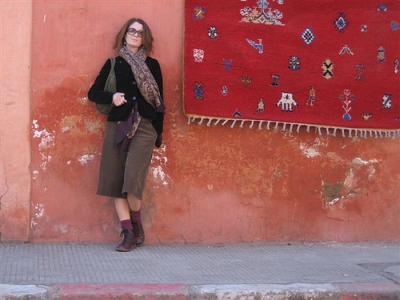}} &
   {\includegraphics[width=0.185\linewidth]{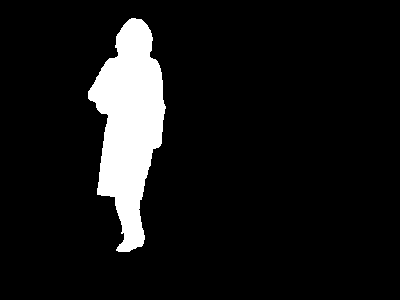}} &
   {\includegraphics[width=0.185\linewidth]{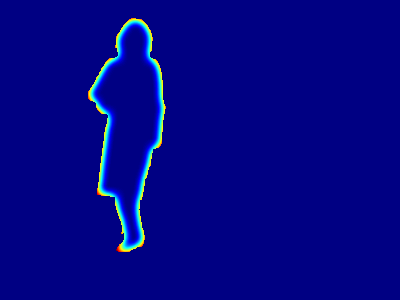}} & 
   {\includegraphics[width=0.185\linewidth]{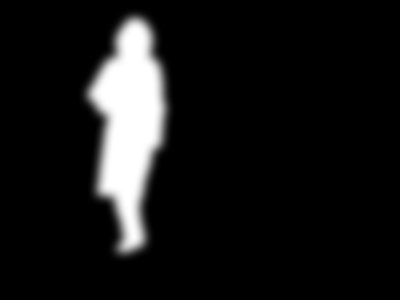}} & {\includegraphics[width=0.185\linewidth]{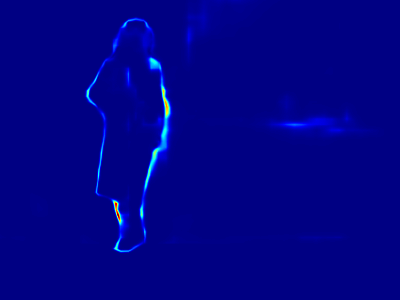}}\\
   \end{tabular}
   \end{center}
   \vspace{-1mm}
\caption{Illustration of boundary distribution smoothing (DBS). From left to right: an input image, its binary ground-truth saliency map, pre-computed uncertainty map, smoothed ground-truth and uncertainty map in the inference stage following \cite{snapshot_uncertainty}.}
\label{fig:uncertainty_pixels}
\end{figure}
Uncertainty measures what a model does not know, which can be systematically categorized into aleatoric uncertainty and epistemic uncertainty~\cite{kendall2017bayesian}.
The aleatoric uncertainty accounts for uncertainty in the data while the epistemic uncertainty represents the model's ignorance of the underlying distribution of the data.
For example, noise in training data can increase the aleatoric uncertainty, while limited data can lead to epistemic uncertainty.

Inspired by the research on uncertainty modeling~\cite{kendall2017uncertainties,kendall2017bayesian,Kendall_2018_CVPR}, our BDS and UATS strategies deal with both aleatoric uncertainty and epistemic uncertainty during training. In particular, with BDS, we target at the aleatoric uncertainty by smoothing out the label distribution,
and UATS tackles the epistemic uncertainty through temperature scaling.
Different from conventional way of using temperature scaling as a post-processing technique \cite{Guo2017OnCO}, and assign a constant temperature to all the samples, we obtain uncertainty-aware dense spatially-variant temperature with each pixel representing uncertainty of current pixel.


\subsection{Boundary Distribution Smoothing}
Following~\cite{rethink_inception}, one could uniformly smooth pixel-wise labels for the whole image, yielding smoothed saliency maps\footnote{For a saliency map $\pmb{y}_i$, we will abuse the notation and denote the smoothed map with $\pmb{y}_i$ as well.} in the form of $\pmb{y}_i \in [0,1]^{h \times w}$ instead of the original $\pmb{y}_i \in \{0,1\}^{h \times w}$. As will be shown empirically in Section \ref{experiments}, this strategy may lead to an under-confident model (see ``M2'' in Table~\ref{tab:base_lines}).

It is generally believed that pixels are not created equally and the uncertainty across the whole image varies from pixel to pixel, where the high-uncertain pixels play a key role \cite{li2017not}. Thus, how to identify the pixels with high uncertainty becomes a key problem in tackling the dense labelling tasks. As observed by Kendall \etal in \cite{kendall2017bayesian}, ``\emph{pixels along the object boundaries are more prone to errors in labelling}''. We perform a similar analysis and report the results in Fig.~\ref{fig:uncertainty_pixels}, where we visualize the pre-computed uncertainty map and the updated uncertainty map following \cite{snapshot_uncertainty}, which shows close connection between object boundaries and prediction uncertainty.


To benefit from this observation, we propose to pre-process the binary ground-truth by considering the underlying uncertainty. Specifically, we intend to assign softened labels in the range of $[0,1]$ to those uncertain region and hard binary labels $\{0,1\}$ to the others following the label smoothing pipeline \cite{rethink_inception}.
Specifically, we gradually smooth those uncertain pixels to produce a continuous saliency map by simply using a Gaussian kernel. As shown in Fig.~\ref{fig:uncertainty_pixels}, the final smoothed map is a continuous map instead of a discrete step map. To produce diverse labels following an adversarial sample generation pipeline \cite{rethink_inception, mixup}, we use multiple Gaussian kernel of different kernel sizes to produce an augmented dataset, which is then our training dataset. 

\subsection{Uncertainty-aware Temperature Scaling}


The BDS strategy focuses on smoothing the supervision signal to better represent the aleatoric uncertainty based on the assumption that labeling error may occurs along object edges.
UATS is proposed to smooth the network output according to the uncertainty of each pixel.

The original Sigmoid function for a binary classification problem is defined as:
\begin{equation}
\label{ori_sigmoid}
    s = \frac{1}{1+\exp({-z})},
\end{equation}
where $z$ and $s$ are the network output and the normalized prediction respectively. 

\begin{figure}[!t]
   \begin{center}
   {\includegraphics[width=0.99\linewidth]{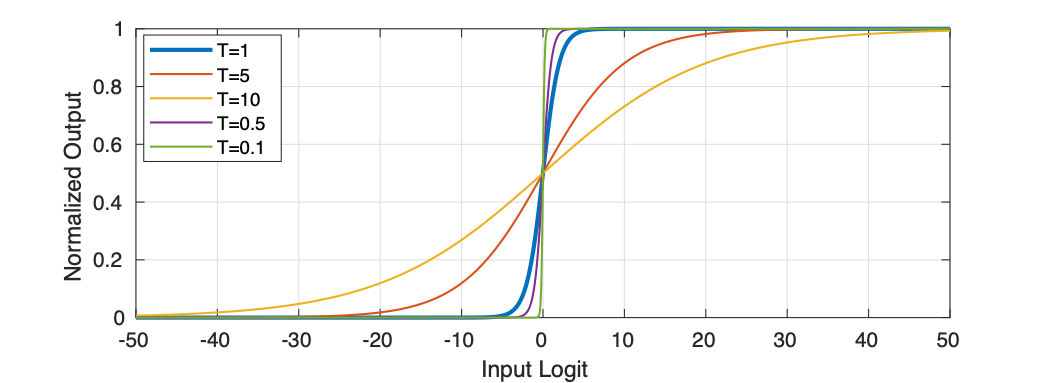}}
   \end{center}
   \vspace{-1mm}
\caption{Illustration of how the relaxed Sigmoid performs with different temperature $T$.}
   \label{fig:sigmoid_temperature}
\end{figure}

Temperature scaling can be introduced to the Sigmoid function, known as the relaxed Sigmoid function, which is defined as:
\begin{equation}
\label{relaxed_sigmoid}
    s' = \frac{1}{1+\exp({-z/T})},
\end{equation}
where $T$ is the temperature and is used to produce softened network prediction $s'$. $T=1$ leads to the original Sigmoid function. $T>1$ produces softened output, and when $T\in(0,1)$, the prediction will collapse to a point mass, as shown in Fig.~\ref{fig:sigmoid_temperature}. 

Taking the uncertainty into consideration, we propose an uncertainty-aware temperature scaling (UATS) method during both training and testing stages. This is different from~\cite{Guo2017OnCO} in at least two aspects: 1) we use sample-dependent and spatial-varing temperature based on uncertainty estimation instead of one uniform temperature for the entire dataset~\cite{Guo2017OnCO}; 2) our temperature is learned during network training, while existing temperature related methods \cite{relaxed_softmax} use pre-defined temperature.

\begin{figure}[!t]
   \begin{center}
   {\includegraphics[width=0.99\linewidth]{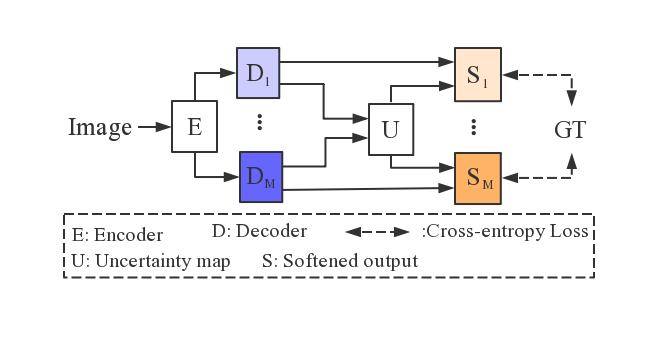}}
   \end{center}
   \vspace{-7mm}
\caption{Training details of our UATS method.
}
   \label{fig:ts_training}
\end{figure}

Specifically, we produce multiple predictions as shown in Fig. \ref{fig:ts_training} following the widely used uncertainty estimation method: M-heads \cite{Rupprecht2016LearningIA}. M-heads is a structured-output generation pipeline, where a shared encoder is to model deterministic feature, and different decoders are used to model stochastic features. We show the M-heads based uncertainty estimation model in Fig.~\ref{fig:ts_training}. Each decoder in the M-heads based framework generate one possible prediction. With $M$ different decoders, we can obtain $M$ predictions. We then compute per-pixel wise uncertainty map $\pmb{U}\in[0,1]$ (variance of those multiple predictions) of the $M$ different predictions, and define temperature $T=\exp(\alpha*\pmb{U})$ during both training and testing, where $\alpha>0$ is a balance parameter.
Thus, for a given image $\pmb{x}_i$, the softened output is achieved as:
\begin{equation}
\label{softten_output}
    \pmb{s}'_i = \frac{1}{1+\exp{(-f(\pmb{x}_i,\Theta)/\exp{(\alpha*\pmb{U}_i)})}}.
\end{equation}
Larger number of $\alpha$ will lead to more smooth prediction, and we set $\alpha=1$ in this paper.

\subsection{Dense Calibration Measure}
\label{sec:dense_calibration_error}
Existing saliency evaluation metrics, including Mean Absolute Error, F-measure, S-measure \cite{fan2017structure} and E-measure \cite{Fan2018Enhanced}, focus only on accuracy of prediction, without considering the gap between accuracy and network confidence. Thus they fail to provide evaluation on how the model is calibrated on a specific dataset.
%
Inspired by~\cite{Guo2017OnCO}, we extend the expected model calibration error to dense prediction task, and define accuracy $\mathrm{acc}$, confidence $\mathrm{conf}$ and dense calibration measure $\mathcal{C}$ of a model on a given dataset.
For model $f(\Theta)$ with parameters $\Theta$ and a given testing set $G=\{\pmb{x}_i,\pmb{y}_i\}^S$ of size $S$, $\mathrm{acc}$ is used to measure the accuracy of the model on $G$. $\mathrm{conf}$ measures how much the model believes in its predictions, and $\mathcal{C}$ measures the calibration error of model $f(\Theta)$ on $G$. 

For an image $\pmb{x}_i$, we define its prediction according to $f(\Theta)$ as $\pmb{s}_i$. Following~\cite{Guo2017OnCO}, we group predictions $\pmb{s}_i$ into $M$ interval bins\footnote{We set $M=12$ in our experiments, with the first and last bin containing predictions of $\pmb{s}_i^{(u,v)} = 0$ and $\pmb{s}_i^{(u,v)} = 1$, respectively.}. The accuracy of each bin is measured as:
\begin{equation}
\label{ori_accuracy}
\mathrm{acc}(B_m)^i=\frac{1}{|B_m|}\sum_{(u,v)\in B_m}\mathbf{1}(g(\pmb{s}_i^{(u,v)})=\pmb{y}_i^{(u,v)}),
\end{equation}
where $B_m$ {are} the samples that fall in the $m$-th interval bin, $|B_m|$ is the cardinality of $B_m$, $(u,v)$ represents coordinate of pixels in $B_m$, $g(.)$ is thresholding operation to transfer gray prediction to binary image. 





As our saliency prediction $\pmb{s}_i$ is a gray scale image, we follow the idea of F-measure, and obtain a binary prediction by thresholding the saliency map $\pmb{s}_i$ in the range of $[0,1]$ with 256 intervals.
Each $g(\pmb{s}_i^{(u,v)})$ lead to one accuracy as Eq. \eqref{ori_accuracy}. With 256 thresholds, we obtain a 256-d vector for accuracy $\mathrm{acc}(B_m)^i$ of each bin. $\mathrm{macc}(B_m)^i$ is defined as mean of $\mathrm{acc}(B_m)^i$. Then, the accuracy $\mathrm{acc}^i$ of image $\pmb{x}_i$ is defined as: $\mathrm{acc}^i = \{\mathrm{macc}(B_1)^i,\cdot,\mathrm{macc}(B_M)^i\}$, which is a $M$ dimensional vector, with each position representing accuracy for a specific bin $B_m$.

The prediction confidence for each pixel represents how much the model trusts its predictions. For an image $\pmb{x}_i$, the average confidence of each bin $B_m$ is then defined as:
\begin{equation}
    \mathrm{conf}(B_m)^i=\frac{1}{|B_m|}\sum_{(u,v)\in B_m} \hat{\pmb{p}}_i^{(u,v)},
\end{equation}
where $\hat{\pmb{p}}_i^{(u,v)}$ is the model confidence at position $(u,v)$, which is defined as:
\begin{equation}
    \hat{\pmb{p}}_i^{(u,v)} = \max\{\pmb{s}_i^{(u,v)},(1-\pmb{s}_i^{(u,v)})\},
\end{equation}



Dense calibration measure $\mathcal{C}$ is the weighted average of the difference between bins' accuracy and confidence. For image $\pmb{x}_i$, we define its dense calibration measure as:
\begin{equation}
    \mathcal{C}^i = \sum_{m=1}^M\frac{|B_m|}{\sum_m |B_m|}|\mathrm{macc}(B_m)^i-\mathrm{conf}(B_m)^i|,
\end{equation}
where $\sum_m |B_m|$ is the number of pixels in $\pmb{x}_i$. A perfectly calibrated model should have $\mathrm{conf}(B_m)^i = \mathrm{macc}(B_m)^i$, thus leads to $\mathcal{C}^i=0$.
For a given testing dataset $G$ and trained model $f(\Theta)$, we define the dense calibration measure of the model on $G$ as: $\mathcal{C}^{G} = \mathrm{mean}\{\mathcal{C}^1,\cdot,\mathcal{C}^S\}$. $\mathcal{C}^{G}=0$ represents the model is perfectly calibrated, and $\mathcal{C}^{G}=1$ indicates poorly calibrated model. In Table \ref{tab:Performance_Comparison}, we show the dense calibration measure of competing methods and ours, which clearly illustrates the effectiveness of our solutions.

\subsection{Implementation Details}
For the boundary distribution smoothing strategy, we apply Gaussian kernel of diverse sizes in the range $(0,5]$ to generate multiple labels for a given image to achieve data augmentation, which is then used as our training dataset.
We build our network on a newly proposed saliency framework (CPD \cite{CPD_Sal} in particular) to test how the proposed strategy can produce well-calibrated saliency model. We set $M=5$, indicating five different decoders, and they share same structure as \cite{CPD_Sal}. We obtain $M$ different predictions during both training and testing, and define spatial-varing temperature of an image based on variance of above multiple predictions according to Eq. \eqref{softten_output}.

We trained our model using Pytorch and integrated both BDS and UATS in our framework, where \enquote{BDS} generates diverse smoothed labels, serving as data augmentation technique, and UATS produces softened output by taking uncertainty of prediction into account. We provide frameworks based on both VGG16 \cite{VGG} and ResNet50 \cite{ResHe2015} backbone following our base model \cite{CPD_Sal}. 
We used the SGD method with momentum 0.9. The base learning rate was initialized as $1\times 10^{-5}$. The whole training took around 13 hours (10 epochs) on a PC with an NVIDIA GeForce RTX GPU.

\section{Experimental Results}
\label{experiments}

\subsection{Setup}
\label{subsec:experimental_setup}
\noindent\textbf{Dataset:} We have evaluated our performance on seven saliency benchmarking datasets. We used 10,553 images from the DUTS dataset~\cite{imagesaliency} for training. The testing datasets include:
1) DUTS testing dataset;
2) ECSSD \cite{Hierarchical:CVPR-2013};
3) DUT \cite{Manifold-Ranking:CVPR-2013};
4) HKU-IS \cite{MDF:CVPR-2015};
5) PASCAL-S \cite{PASCALS};
6) THUR \cite{THUR};
7) MSRA-B testing dataset \cite{DRFI:CVPR-2013}.

\noindent\textbf{Competing methods:} We have compared our methods against twelve fully supervised deep salient object detection models as shown in Table \ref{tab:BenchmarkResults}.

\noindent\textbf{Evaluation metrics:} Four evaluation metrics are used for performance evaluation, including two widely used (mean absolute error (MAE $\mathcal{M}$), F-measure ($F_{\beta}$)), one newly proposed (S-measure $S\alpha$ ~\cite{fan2017structure}), and our proposed dense calibration measure ($\mathcal{C}$). 

\begin{table*}[t!]
  \centering
  \scriptsize
  \renewcommand{\arraystretch}{1.3}
  \renewcommand{\tabcolsep}{1.3mm}
  \caption{Benchmarking results of competing SOD models on seven datasets.
  $\uparrow \& \downarrow$ denote larger and smaller is better, respectively.
  }\label{tab:BenchmarkResults}
  \begin{tabular}{lr|ccccccccc|ccccc|cc}
  \hline
  &  &\multicolumn{9}{c|}{VGG16 backbone}&\multicolumn{5}{c|}{ResNet50 backbone}&\multicolumn{2}{c}{Others} \\
    & Metrics &
   Amulet  & DSS  & PiCANet & RAS    & NLDF &
   MSNet & CPD & AFNet & Ours  & PiCANet& CPD & DGRL & SRM & Ours &  BASNet & PAGRN\\
   &  & \cite{Amulet}        & \cite{ChengCVPR17}       & \cite{picanet}          & \cite{ras_reverse}              & \cite{wangiccv17} &
        \cite{MSNet_Sal}   & \cite{CPD_Sal}                 & \cite{AFNet_Sal}  & 
        & \cite{picanet}   & \cite{CPD_Sal} &\cite{Wang_2018_CVPR}      & \cite{wangiccv17} &   &  \cite{BASNet_Sal}     & \cite{prpgressive_attention}\\
  \hline
  \multirow{4}{*}{\begin{sideways}\textit{DUTS}\cite{imagesaliency}\end{sideways}}
    & $S_{\alpha}\uparrow$    & .7928 & .7889 & .8423 & .7918 & .8162 & .8617 & .8668 & .8671 & \textbf{.8865} & .8514 & .8690 & .8460 & .8358 & \textbf{.8761} & .8657 & .8385 \\
    & $F_{\beta}\uparrow$     & .6893 & .7286 & .7565 & .7410 & .7567 & .7917 & .8246 & .8123 & \textbf{.8528} & .7662 & .8208 & .7898 & .7655 & \textbf{.8302} & .8226& .7781   \\
    & $\mathcal{C}\downarrow$ & .0536 & .0597 & .0357 & .0447 & .0473 & .0323 & .0351 & .0326 &\textbf{.0253} & .0301 & .0322 & .0452 & .0363 & \textbf{.0223}  & .0410& .0365  \\
    & $\mathcal{M}\downarrow$ & .0916 & .0749 & .0621 & .0746 & .0652 & .0490 & .0428 & .0457 & \textbf{.0352} & .0581 & .0434 & .0512 & .0578 & \textbf{.0380}  & .0476 & .0555 \\ \hline
  \multirow{4}{*}{\begin{sideways}\textit{ECSSD}\cite{Hierarchical:CVPR-2013}\end{sideways}}
    & $S_{\alpha}\uparrow$    & .8905 & .8236 & .8984 & .8211 & .8697 & .9048 & .9046 & .9074 & \textbf{.9205} & .9061 & .9129 & .9019 & .8907 & \textbf{.9264} & .9104 & .8825 \\
    & $F_{\beta}\uparrow$     & .8704 & .8344 & .8719 & .8372 & .8714 & .8856 & .9076 & .9008 & \textbf{.9225} & .8794 & .9093 & .8978 & .8809 & \textbf{.9297} & .9128& .8718   \\
    & $\mathcal{C}\downarrow$ & .0392 & .0721 & .0284 & .0429 & .0476 & .0305 &.0358 & .0320 & \textbf{.0251} & .0248 & .0292 & .0383 & .0364 & \textbf{.0214} & .0336& .0414   \\
    & $\mathcal{M}\downarrow$ & .0608 & .0895 & .0543 & .0899 & .0655 & .0479 & .0434 & .0450 & \textbf{.0358} & .0519 & .0397 & .0447 & .0566 & \textbf{.0311} & .0399& .0644  \\ \hline
  \multirow{4}{*}{\begin{sideways}\textit{DUT}\cite{Manifold-Ranking:CVPR-2013}\end{sideways}}
    & $S_{\alpha}\uparrow$    & .7805 & .7441 & .8169 & .7620 & .7704 & .8093 & .8177 & .8263 & \textbf{.8456} & .8237 & .8248 & .8097 & .7977 & \textbf{.8327} & .8362& .7751   \\
    & $F_{\beta}\uparrow$     & .6670 & .6618 & .7105 & .6897 & .6825 & .7095 & .7385 & .7425 & \textbf{.7812} & .7158 & .7417 & .7264 & .6970 & \textbf{.7589}  & .7668& .6754  \\
    & $\mathcal{C}\downarrow$ & .0639 & .0720 & .0432 & .0475 & .0603 & .0481 & .0487 & .0443 & \textbf{.0374} & .0388 & .0438 & .0574 & .0464 & \textbf{.0362}  & .0496& .0540  \\
    & $\mathcal{M}\downarrow$ & .0976 & .0867 & .0722 & .0793 & .0796 & .0636 & .0567 & .0574 & \textbf{.0464} & .0693 & .0560 & .0632 & .0694 & \textbf{.0501}  & .0565& .0709 \\ \hline
  \multirow{4}{*}{\begin{sideways}\textit{HKU-IS}\cite{MDF:CVPR-2015}\end{sideways}}
    & $S_{\alpha}\uparrow$    & .8834 & .8473 & .8949 & .8394 & .8787 & .9065 & .9039 & .9053 & \textbf{.9176} & .8948 & .9055 & 8968 & .8870 & \textbf{.9232}  & .9089& .8872  \\
    & $F_{\beta}\uparrow$     & .8449 & .8452 & .8543 & .8484 & .8711 & .8780 & .8948 & .8877 & \textbf{.9043} & .8517 & .8921 & .8844 & .8664 & \textbf{.9166}  & .9025& .8638  \\
    & $\mathcal{C}\downarrow$ & .0278 & .0453 & .0222 & .0299 & .0324 & .0226 & .0259 & .0233 & \textbf{.0175} & .0205 & .0235 & .0304 & .0254 & \textbf{.0154}  & .0257 & .0262 \\
    & $\mathcal{M}\downarrow$ & .0519 & .0607 & .0464 & .0627 & .0477 & .0387 & .0333 & .0358 & \textbf{.0289} & .0479 & .0342 & .0374 & .0459 & \textbf{.0259}  & .0322 & .0475 \\ \hline
   \multirow{4}{*}{\begin{sideways}\textit{PASCAL-S}\cite{PASCALS}\end{sideways}}
    & $S_{\alpha}\uparrow$    & .7937 & .7124 & .7877 & .6940 & .7559 & .7944 & .7860 & .7968 & \textbf{.8053} & .7898 & .7893 & .7959 & .7816 & \textbf{.8082} & .7498 & .7846 \\
    & $F_{\beta}\uparrow$     & .8064 & .7568 & .7985 & .7546 & .7933 & .8129 & .8220 & .8241 & \textbf{.8452} & .7945 & .8150 & .8289 & .8026 & \textbf{.8422} & .8212& .7656  \\
    & $\mathcal{C}\downarrow$ & .0953 & .1534 & .0964 & .1423 & .1233 & .0999 & .1119 & .0998 & \textbf{.0973} & .0934 & .1066 & .1059 & .1035 & \textbf{.0942} & .1135& .1257  \\
    & $\mathcal{M}\downarrow$ & .1292 & .1720 & .1284 & .1812 & .1454 & .1193 & .1215 & .1155 & \textbf{.1092} & .1284 & .1202 & .1150 & .1313 & \textbf{.1078}  & .1217& .1516  \\ \hline
   \multirow{4}{*}{\begin{sideways}\textit{THUR}\cite{THUR}\end{sideways}}
    & $S_{\alpha}\uparrow$    & .7965 & .7720 & .8181 & .7798 & .8008 & .8188 & .8311 & .8251 & \textbf{.8442} & .8233& .8345  & .8162 & .8179 & \textbf{.8466}  & .8232& .8304  \\
    & $F_{\beta}\uparrow$     & .6865 & .6875 & .7098 & .7003 & .7111 & .7177 & .7498 & .7327 & \textbf{.7689} & .7133 & .7504 & .7271 & .7201 & \textbf{.7598}  & .7366& .7395  \\
    & $\mathcal{C}\downarrow$ & .0657 & .0727 & .0627 & .0701 & .0620 & .0625 & .0603 & .0587 &\textbf{.0471} & .0536& .0576 & .0709 & .0542 & \textbf{.0440}  & .0662 & .0486 \\ 
    & $\mathcal{M}\downarrow$ & .0936 & .0893 & .0836 & .0833 & .0805 & .0794 & .0935 & .0724 & \textbf{.0613} & .0816& .0680  & .0774 & .0769 & \textbf{.0612} & .0734 & .0704 \\ \hline
    \multirow{4}{*}{\begin{sideways}\textit{MSRA-B}\cite{DRFI:CVPR-2013}\end{sideways}}
    & $S_{\alpha}\uparrow$    & - & .8660 & .9055 & .8730 & .9100 & - & .9079 & .9062 & \textbf{.9160} & .9100& .9183  & .8999 & .8415 & \textbf{.9256}  & .9102& -  \\
    & $F_{\beta}\uparrow$     & - & .8614 & .8666 & .8713 & .8694 & - & .8914 & .8830 & \textbf{.8974} & .8694 & .8999 & .8866 & .8053 & \textbf{.9038}  & .9012& -  \\
    & $\mathcal{C}\downarrow$ & - & .0463 & .0293 & .0448 & .0342 & - & .0316 & .0310 &\textbf{.0298} & .0259& .0264 & .0371 & .0467 & \textbf{.0223}  & .0279 & - \\
    & $\mathcal{M}\downarrow$ & - & .0596 & .0501 & .0555 & .0497 & -  & .0385 & .0431 & \textbf{.0301}& .0497& .0358  & .0416 & .0700 & \textbf{.0307} & . 0398& - \\
  \hline
  \end{tabular}
  \label{tab:Performance_Comparison}
 \end{table*}
\begin{figure*}[!htp]
   \begin{center}
   {\includegraphics[width=0.245\linewidth]{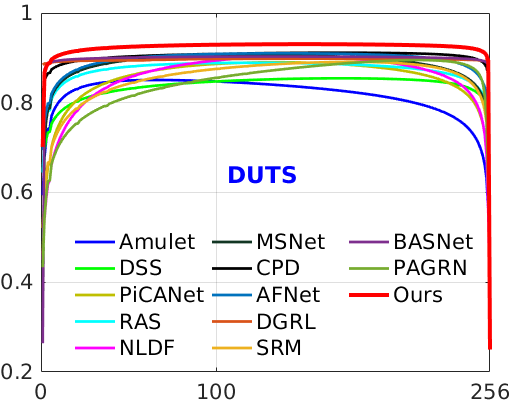}}
   {\includegraphics[width=0.245\linewidth]{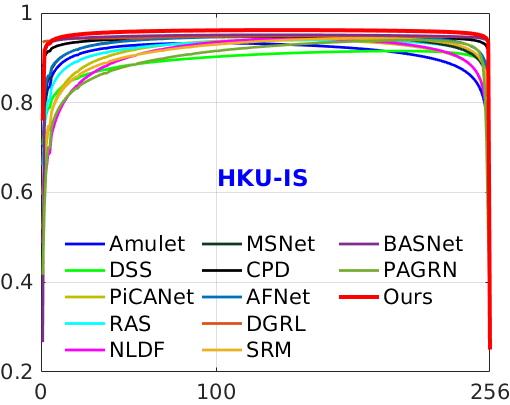}}
   {\includegraphics[width=0.245\linewidth]{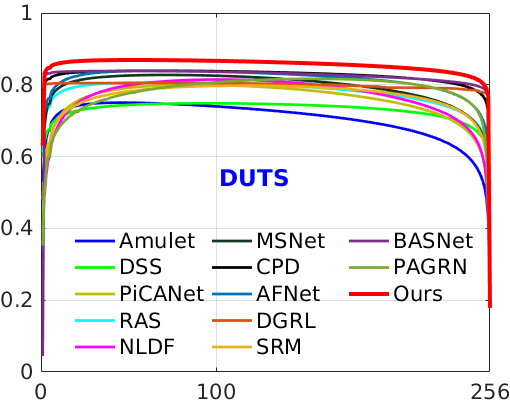}}
   {\includegraphics[width=0.245\linewidth]{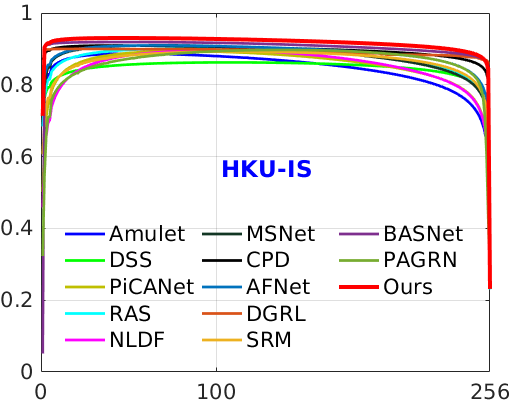}}
   \end{center}
   \vspace{-2mm}
   \caption{E-measure and F-measure curves on two testing dataset. First two figures: E-measure. Last two figures: F-measure.}
   \label{fig:pr_curve}
\end{figure*}

\begin{figure*}[!t]
   \begin{center}
   \begin{tabular}{ c@{ } c@{ } c@{ } c@{ }  c@{ }  c@{ } c@{ } c@{ } c@{ }}
   {\includegraphics[width=0.105\linewidth]{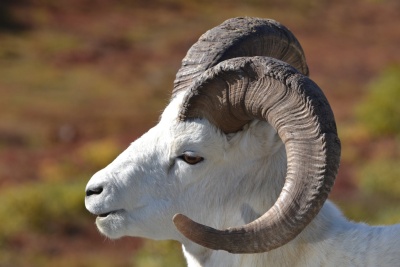}}&
   {\includegraphics[width=0.105\linewidth]{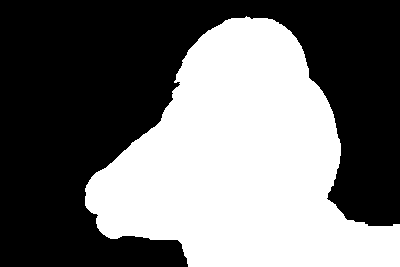}}&
   {\includegraphics[width=0.105\linewidth]{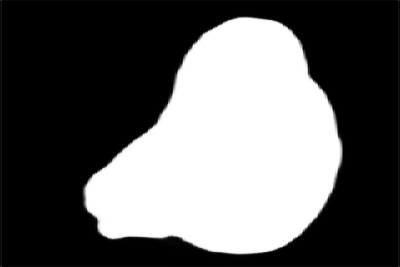}}&
   {\includegraphics[width=0.105\linewidth]{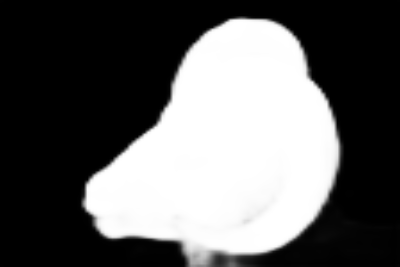}}&
   {\includegraphics[width=0.105\linewidth]{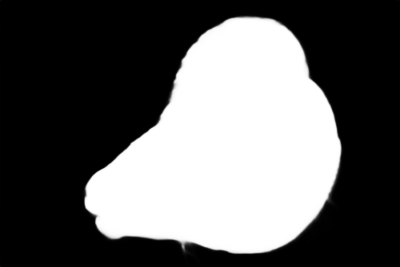}}&
   {\includegraphics[width=0.105\linewidth]{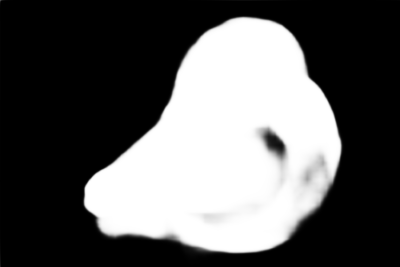}}&
   {\includegraphics[width=0.105\linewidth]{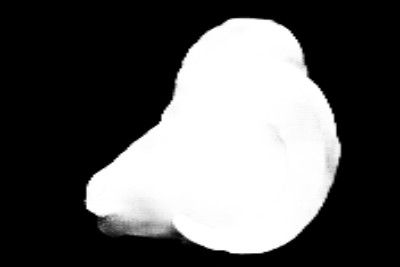}}&
   {\includegraphics[width=0.105\linewidth]{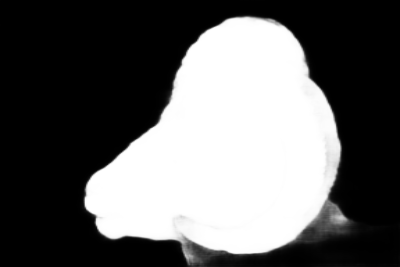}}&
   {\includegraphics[width=0.105\linewidth]{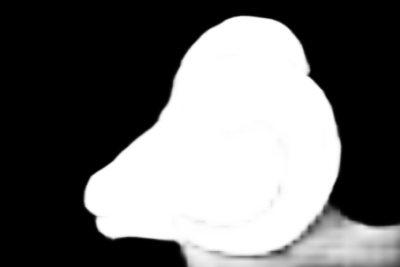}}\\
      {\includegraphics[width=0.105\linewidth]{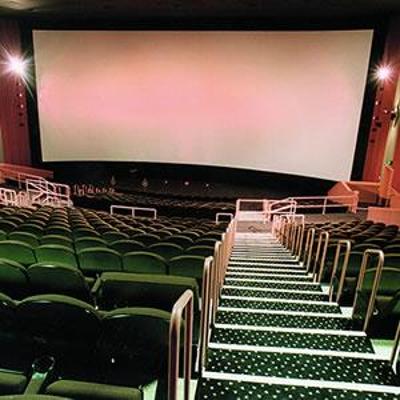}}&
   {\includegraphics[width=0.105\linewidth]{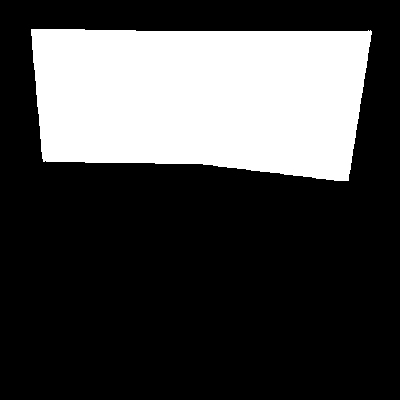}}&
   {\includegraphics[width=0.105\linewidth]{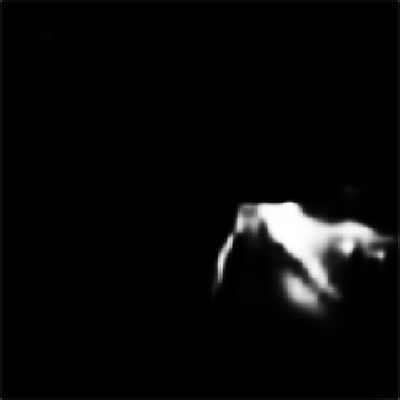}}&
   {\includegraphics[width=0.105\linewidth]{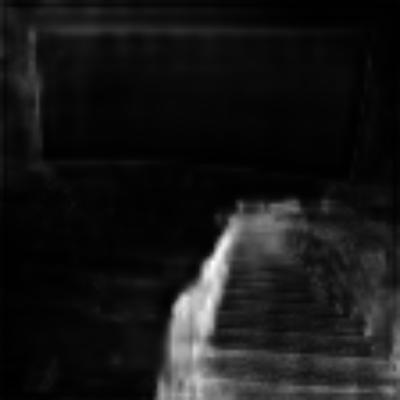}}&
   {\includegraphics[width=0.105\linewidth]{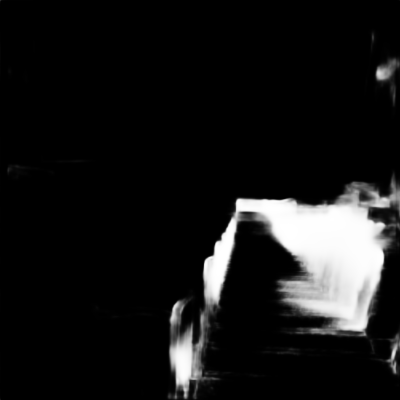}}&
   {\includegraphics[width=0.105\linewidth]{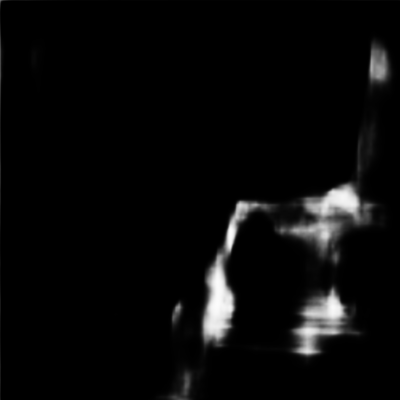}}&
   {\includegraphics[width=0.105\linewidth]{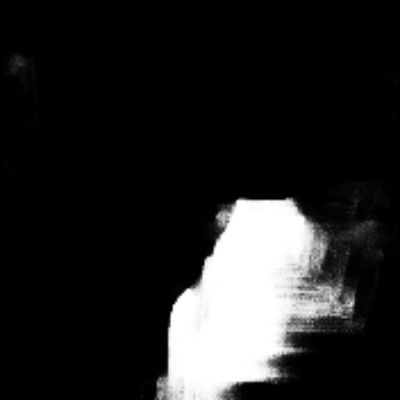}}&
   {\includegraphics[width=0.105\linewidth]{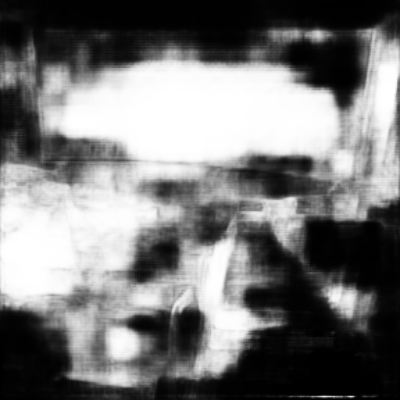}}&
   {\includegraphics[width=0.105\linewidth]{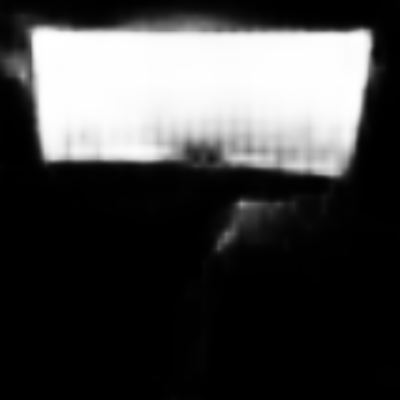}}\\
   {\includegraphics[width=0.105\linewidth]{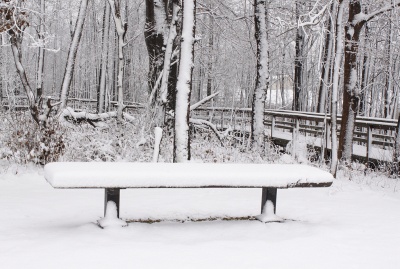}}&
   {\includegraphics[width=0.105\linewidth]{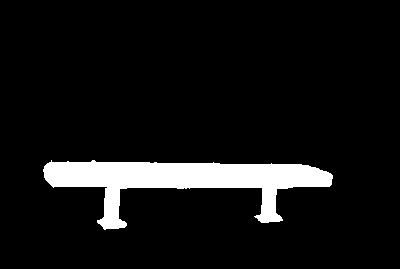}}&
   {\includegraphics[width=0.105\linewidth]{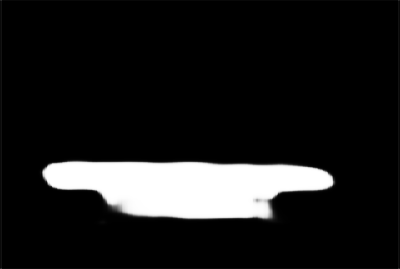}}&
   {\includegraphics[width=0.105\linewidth]{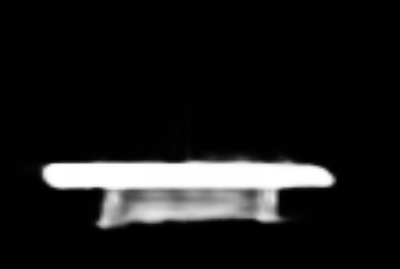}}&
   {\includegraphics[width=0.105\linewidth]{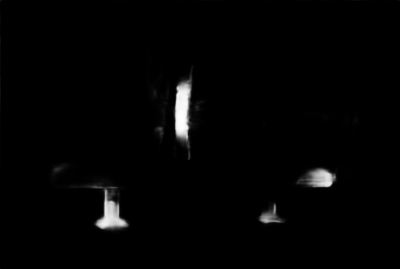}}&
   {\includegraphics[width=0.105\linewidth]{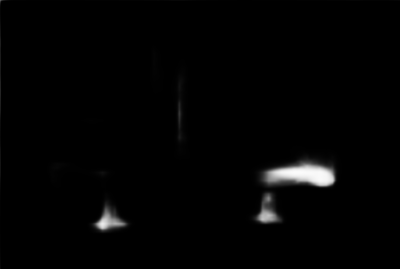}}&
   {\includegraphics[width=0.105\linewidth]{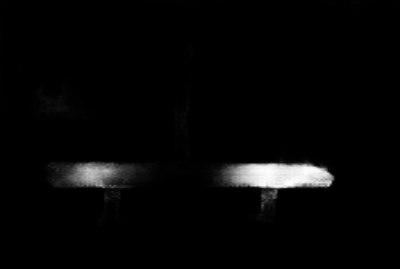}}&
   {\includegraphics[width=0.105\linewidth]{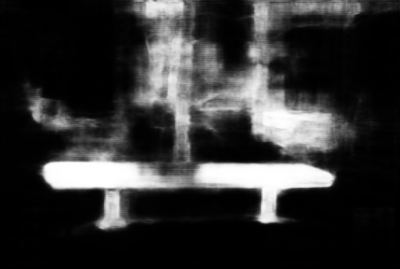}}&
   {\includegraphics[width=0.105\linewidth]{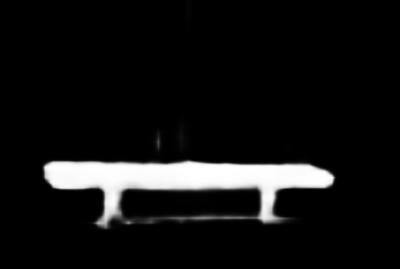}}\\
    {\includegraphics[width=0.105\linewidth]{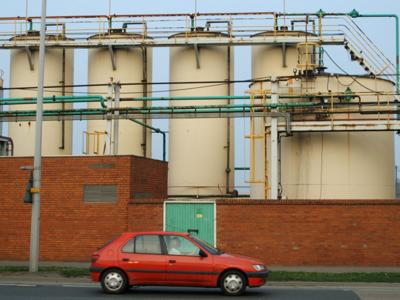}}&
   {\includegraphics[width=0.105\linewidth]{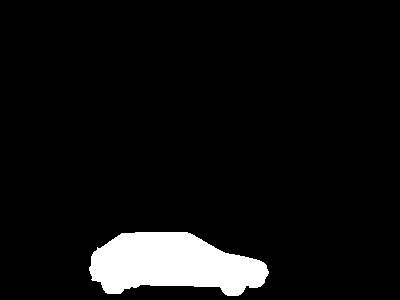}}&
   {\includegraphics[width=0.105\linewidth]{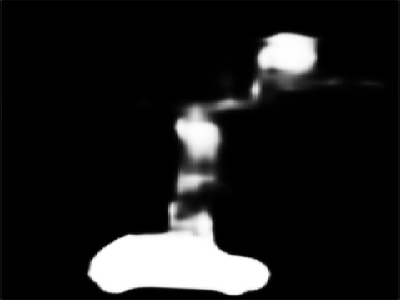}}&
   {\includegraphics[width=0.105\linewidth]{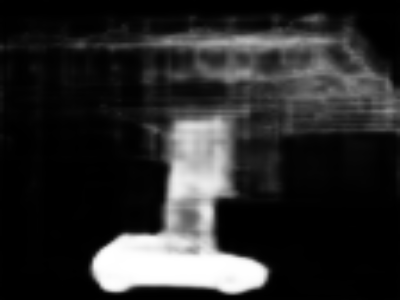}}&
   {\includegraphics[width=0.105\linewidth]{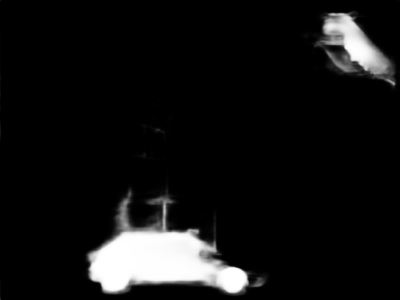}}&
   {\includegraphics[width=0.105\linewidth]{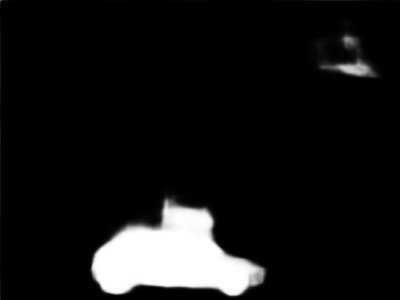}}&
   {\includegraphics[width=0.105\linewidth]{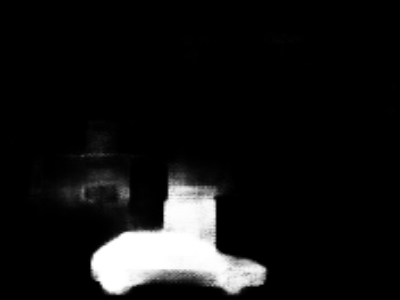}}&
   {\includegraphics[width=0.105\linewidth]{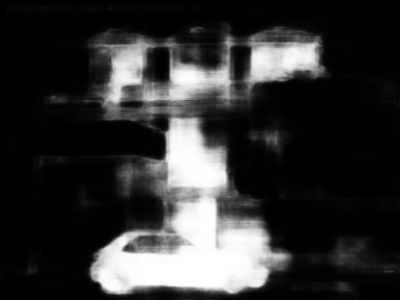}}&
   {\includegraphics[width=0.105\linewidth]{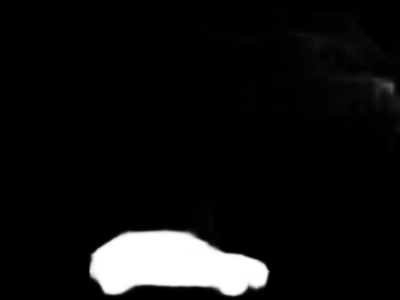}}\\
   \footnotesize{Image} & \footnotesize{GT} & \footnotesize{CPD} & \footnotesize{PiCANet} & \footnotesize{RAS} & \footnotesize{DSS} & \footnotesize{NLDF} & \footnotesize{Amulet} & \footnotesize{Ours}\\
   \end{tabular}
   \end{center}
   \vspace{-2mm}
\caption{\small Visual comparison between our method and other competing methods.
}
   \label{fig:saliency_compare}
\end{figure*}

\begin{table}[t!]
  \centering
  \scriptsize
  \renewcommand{\arraystretch}{1.3}
  \renewcommand{\tabcolsep}{1.6mm}
  \caption{Performance of experiments in the ablation study section.
  }\label{tab:base_lines}
  \begin{tabular}{lr|ccccccc}
  \hline
    & Metrics &
   M0  & Base  & M1 & M2    & M3 &
   M4 & M5 \\
  \hline
  \multirow{4}{*}{\begin{sideways}\textit{DUTS}\cite{imagesaliency}\end{sideways}}
    & $S_{\alpha}\uparrow$    & .8865 & .8668 & .8732 & .8012  & .8790 & .8341 & .8712   \\
    & $F_{\beta}\uparrow$     & .8528 & .8246 & .8470 & .7823  & .8510 & .8165 & .8367   \\
    & $\mathcal{C}\downarrow$ & .0253 & .0351 & .0302 & .0510  & .0289 & .0311 & .0296  \\
    & $\mathcal{M}\downarrow$ & .0352 & .0428 & .0415 & .0555  & .0411 & .0485 & .0434   \\ \hline
  \multirow{4}{*}{\begin{sideways}\textit{ECSSD}\cite{Hierarchical:CVPR-2013}\end{sideways}}
    & $S_{\alpha}\uparrow$    & .9205 & .9046 & .9134 & .8704  & .9134 & .8838 & .9028   \\
    & $F_{\beta}\uparrow$     & .9225 & .9076 & .9170 & .8572  & .9189 & .8857 & .8982   \\
    & $\mathcal{C}\downarrow$ & .0251 & .0358 & .0310 & .0573  & .0301 & .0312 & .0259   \\
    & $\mathcal{M}\downarrow$ & .0358 & .0434 & .0416 & .0631  & .0451 & .0487 & .0460   \\ \hline
  \multirow{4}{*}{\begin{sideways}\textit{DUT}\cite{Manifold-Ranking:CVPR-2013}\end{sideways}}
    & $S_{\alpha}\uparrow$    & .8456 & .8263 & .8302 & .7521  & .8389 & . 8124& .8270  \\
    & $F_{\beta}\uparrow$     & .7812 & .7425 & .7527 & .7192  & .7676 & .7344 & .7442   \\
    & $\mathcal{C}\downarrow$ & .0374 & .0443 & .0411 & .0595  & .0418 & .0418 & .0419   \\
    & $\mathcal{M}\downarrow$ & .0464 & .0574 & .0546 & .0670  & .0556 & .0594 & .0587   \\ \hline
  \multirow{4}{*}{\begin{sideways}\textit{HKU-IS}\cite{MDF:CVPR-2015}\end{sideways}}
    & $S_{\alpha}\uparrow$    & .9176 & .9039 & .9102 & .8603  & .9132 & .9014 & .9092  \\
    & $F_{\beta}\uparrow$     & .9043 & .8948 & .9028 & .8252  & .9019 & .8803 & .8974   \\
    & $\mathcal{C}\downarrow$ & .0175 & .0259 & .0210 & .0417  & .0213 & .0225 & .0230  \\
    & $\mathcal{M}\downarrow$ & .0289 & .0333 & .0317 & .0551  & .0346 & .0303 & .0311  \\ \hline
   \multirow{4}{*}{\begin{sideways}\textit{PASCAL-S}\cite{PASCALS}\end{sideways}}
    & $S_{\alpha}\uparrow$    & .8053 & .7860 & .7901 & .7607  & .7931 & .7691 & .7891  \\
    & $F_{\beta}\uparrow$     & .8452 & .8220 & .8310 & .8177  & .8303 & .8043 & .8190   \\
    & $\mathcal{C}\downarrow$ & .0973 & .1119 & .1102 & .1312  & .1052 & .1034 & .1111  \\
    & $\mathcal{M}\downarrow$ & .1092 & .1215 & .1176 & .1395  & .1118 & .1285 & .1209   \\ \hline
   \multirow{4}{*}{\begin{sideways}\textit{THUR}\cite{THUR}\end{sideways}}
    & $S_{\alpha}\uparrow$    & .8442 & .8311 & .8349 & .7815  & .8314 & .8118 & .8267   \\
    & $F_{\beta}\uparrow$     & .7689 & .7498 & .7544 & .7051  & .7529 & .7238 & .7509 \\
    & $\mathcal{C}\downarrow$ & .0471 & .0603 & .0562 & .0816  & .0519 & .0578 & .0586  \\
    & $\mathcal{M}\downarrow$ & .0613 & .0935 & .0668 & .1143  & .0646 & .0666 & .0698   \\ \hline
    \multirow{4}{*}{\begin{sideways}\textit{MSRA-B}\cite{DRFI:CVPR-2013}\end{sideways}}
    & $S_{\alpha}\uparrow$    & .9160 & .9079 & .9110 & .8716  & .9096 & 8756 & .9000  \\
    & $F_{\beta}\uparrow$     & .8974 & .8914 & .8986 & .8385  & .8853 & .8612 & .8807   \\
    & $\mathcal{C}\downarrow$ & .0298 & .0316 & .0300 & .0591  & .0308 & .0315 & .0312  \\
    & $\mathcal{M}\downarrow$ & .0301 & .0385 & .0338 & .0608  & .0411 & .0471 & .0426  \\
  \hline
  \end{tabular}
 \end{table}

\subsection{Comparison with State-of-the-Art}
\noindent\textbf{Comparisons on Calibration Measure:}
We computed the dense calibration measure $\mathcal{C}$ of competing methods and ours, and show results in Table \ref{tab:BenchmarkResults}.
We discovered that the proposed solution achieves consistently the smallest $\mathcal{C}$ measure with both ResNet50 backbone and VGG16 backbone. Further, we find that for those SOD models with both VGG and ResNet backbones (PiCANet \cite{picanet} and CPD \cite{CPD_Sal} in particular), their ResNet50 based models achieves smaller dense calibration measure, indicating better generalization ability for the ResNet50 based models.

Moreover, through carefully analysing the $\mathcal{C}$ measure for competing methods on all the testing datset, we find that the proposed dense calibration measure do not necessarily consistent with other metrics. For example, S-measure of RAS \cite{ras_reverse} on ECSSD dataset is worse than that of NLDF, while we observe smaller dense calibration measure for RAS. This phenomenon indicates that dense calibration measure can discover other attributes of deep SOD models, and together with existing eveluation metrics to provide more comprehensive evaluation of a given model.


\noindent\textbf{Quantitative Comparison:} We compared our method with state-of-the-art SOD methods, and the performance is reported in Table~\ref{tab:BenchmarkResults} and Fig.~\ref{fig:pr_curve}, where \enquote{Ours} represents result of our model trained with M-heads based on the augmented dataset through label smoothing. 
We observe consistent performance improvement, around $2\%$ improvement of S-measure and F-measure, as well as around $1\%$ decease of MAE. As mentioned above, we build our network on CPD \cite{CPD_Sal}, and we add four extra decoder to CPD network, and train the model with augmented data through label smoothing. We notice that the proposed solution not only improve network performance, but also lead to a better calibrated model.
In Fig.~\ref{fig:pr_curve}, we show E-measure and F-measure curves on two datasets (we have both curves on seven testing datasets, and only show two of them due to page limit). We observe that although CPD has achieved very good performance with E-measure and F-measure on top of the curves, our proposed solutions can further boost its performance, and achieve the best performance compared with competing methods.


\noindent\textbf{Qualitative Comparisons:}
In Fig.~\ref{fig:saliency_compare}, we presented five visual comparisons between our methods and competing methods, where our methods produce comparable or best performance. The salient object in the first row is large, where part of the salient foreground is in a shadow area and shares similar appearance with the background. Most of the existing deep models fail to detect that region. The proposed methods can achieve better results with the above hard region highlighted. The salient object in the second row expands to a large region. Most of the competing methods incorrectly predict salient object as background. While the proposed two solutions can preserve more salient foreground. The salient object in the third image appears quite similar to the background, which makes almost all of the competing deep models fail to distinguish the salient foreground from the background, especially for the legs part of the salient object, while our methods produce better saliency maps, with most of the background removed. The background in the fourth image is quite complex and similar to the foreground, competing deep models usually fail to discriminate salient objects from the clustered background, while our method produces nearly clear salient maps. 

\subsection{Ablation Studies}
We carried out two experiments (\enquote{M1}, and \enquote{M2}) to analyze label smoothing and another three experiments (\enquote{M3}, \enquote{M4} and \enquote{M5}) to illustrate the effectiveness of the proposed uncertainty-aware temperature scaling technique.
Performance of all the experiments in this section is shown in Table \ref{tab:base_lines}, where \enquote{M0} is our final performance, and \enquote{Base} represents performance of the base model (CPD \cite{CPD_Sal} in particular).

\noindent\textbf{Base Model $+$ BDS:}
We use diverse scales of Gaussian kernel to achieve augmented dataset, and train the base model \cite{CPD_Sal} with the augmented dataset. We show performance of this experiment as \enquote{M1}. Compared with performance of our base model \enquote{Base}, we observe consistent improved performance and with lower $\mathcal{C}$ measure in \enquote{M1}, which can be explained  from at least two aspects: 1) through label smoothing, we achieve data augmentation; 2) the smoothed data can be seen as adversarial samples, and training on it can improve the network generalization ability.

\noindent\textbf{Base Model $+$ Uniform Label Smoothing:}
Conventional way of using label smoothing \cite{rethink_inception} is for image-level classification task, where they uniformly smooth the one-hot label vector to produce adversarial samples. Following this basic setting, we use uniform smoothing instead, and assign pixels along salient object edges a uniform saliency value, and the performance is reported as \enquote{M2}. We find inferior performance of \enquote{M2} compared with the base model. The main reason is that during uniform label smoothing, we introduce noise to the network. Although uniform label smoothing works well in the image-level classification tasks, it is more appropriate to use a Gaussian smoothing for our dense prediction task.



\noindent\textbf{Base Model $+$ UATS:}
With the base model, we embed the proposed UATS (uncertainty-aware temperature scaling) module to the network, and show the performance as \enquote{M3}. We observe that UATS not only consistently improves network performance, but also achieves smaller dense calibration measure, leading to a well-calibrated model.

\noindent\textbf{Base Model $+$ Uniform TS as Post-processing:}
Guo \etal \cite{Guo2017OnCO} introduced temperature scaling as post-processing for model calibration, where a uniform temperature is assign to the entire dataset to produce softened output. Following this setting, we define spatial-sample-independent uniform temperature $T=2$ for the entrie dataset, and show the performance as \enquote{M4}. We notice that \enquote{M4} can indeed reduce the dense calibration measure, while it also cause inferior performance compared with the base model. 
This experiment indicates that although TS as post-processing works well in image-level classification, it may not be the right solution for dense prediction task, where spatial-varying temperature is more desirable. 

\noindent\textbf{Base Model $+$ Adaptive TS as Post-processing:}
\enquote{M4} assigns constant temperature for the entire dataset without considering the uniqueness of each sample. Based on our observation as shown in Fig.~\ref{fig:uncertainty_pixels}, high-uncertain pixels usually exist along object edges. We compute image edge map $e$, and define temperature $T=\exp(e)$, which assigns high temperature to pixels along object edges, and constant temperature ($T=1$) to pixels inside object regions. We show the performance as \enquote{M5}. Compared with \enquote{Base}, we observe improved accuracy and decreased $\mathcal{C}$ measure, which indicates the effectiveness of spatially-varying temperature. Meanwhile, our result in \enquote{M0} with uncertainty maps from the network achieves even better performance, which further proves the benefit of learning uncertainty maps.

\vspace{-2mm}
\section{Conclusion}
\vspace{-2mm}
In this paper, we addressed the confidence calibration issue for deep SOD, \ie, the gap between network accuracy and prediction confidence. First, we showed that state-of-the-art SOD networks are prone to producing overconfident predictions. Then we proposed two strategies to resolve the overconfidence issue with SOD networks, namely the boundary distribution smoothing strategy (BDS) and the uncertainty-aware temperature scaling strategy (UATS). 
BDS addresses the overconfidence issue by applying Gaussian kernels to blur the ground-truth labels in order to produce a continuous supervision signal.
UATS assigns temperature to different pixels by considering their uncertainty adpatively. 
Experimental results on seven benchmark datasets proved the effectiveness of our solutions. 

In the future, we plan to extend the current binary dense prediction framework to other dense tasks, such as monocular depth estimation~\cite{Zhong2018ECCV}, stereo matching~\cite{zhong2018open,cheng2020hierarchical}, optical flow~\cite{zhong2019unsupervised,wang2020displacement}, semantic segmentation \cite{FCN} and instance-level object segmentation \cite{MASK-R-CNN}. 

{\small
\bibliographystyle{ieee_fullname}
\bibliography{SaliencyBib}
}

\end{document}